\documentclass[review]{elsarticle}

\usepackage{times}
\usepackage{url}
\usepackage{latexsym}
\usepackage{graphicx}
\usepackage{enumerate}
\usepackage{subfigure}

\usepackage{makecell}
\usepackage{multirow}
\usepackage[small]{caption}
\usepackage{amsmath,amsfonts}
\usepackage{amsthm}
\usepackage{booktabs}
\usepackage{algorithm}
\usepackage{algorithmic}
\usepackage{bbm}
\usepackage[x11names]{xcolor}
\usepackage{verbatim}
\usepackage[T1]{fontenc}
\usepackage[utf8]{inputenc}
\usepackage{float}
\usepackage{url}
\usepackage{hyperref}
\hypersetup{hidelinks,
	colorlinks=true,
	allcolors=black,
	pdfstartview=Fit,
	breaklinks=true}

\journal{}

\begin{document}

\begin{frontmatter}

\title{Neural Topic Modeling with Deep Mutual Information Estimation}


\author[mymainaddress]{Kang Xu}
\ead{kxu@njupt.edu.cn}

\author[mysecondaryaddress]{Xiaoqiu Lu}
\ead{luxq0823@163.com}

\author[myfouraddress]{Yuan-fang Li}
\ead{yuanfang.li@monash.edu}

\author[mysecondaryaddress]{Tongtong Wu}
\ead{wutong8023@seu.edu.cn}

\author[mysecondaryaddress]{Guilin Qi\corref{mycorrespondingauthor}}
\cortext[mycorrespondingauthor]{This is to indicate the corresponding author.}
\ead{gqi@seu.edu.cn}

\author[mymainaddress]{Ning Ye}
\ead{yening@njupt.edu.cn}

\author[myfifthaddress]{Dong Wang}
\ead{wangdong19@mails.tsinghua.edu.cn}

\author[myfifthaddress]{Zheng Zhou}
\ead{zhouz@mail.tsinghua.edu.cn}

\address[mymainaddress]{School of Computer Science, Nanjing University of Posts and Telecommunications, China}


\address[mysecondaryaddress]{School of Computer Science and Engineering, Southeast University,China}

\address[myfouraddress]{Faculty of Information Technology, Monash University, Australia}

\address[myfifthaddress]{Department of Earth System Science, Tsinghua University, Beijing 100084, China}

\begin{abstract}
The emerging neural topic models make topic modeling more easily adaptable and extendable in unsupervised text mining. However, the existing neural topic models is difficult to retain representative information of the documents within the learnt topic representation. In this paper, we propose a neural topic model which incorporates deep mutual information estimation, i.e., Neural Topic Modeling with Deep Mutual Information Estimation(NTM-DMIE). NTM-DMIE is a neural network method for topic learning which maximizes the mutual information between the input documents and their latent topic representation. To learn robust topic representation, we incorporate the discriminator to discriminate negative examples and positive examples via adversarial learning. Moreover, we use both global and local mutual information to preserve the rich information of the input documents in the topic representation. We evaluate NTM-DMIE on several metrics, including accuracy of text clustering, with topic representation, topic uniqueness and topic coherence. Compared to the existing methods, the experimental results show that NTM-DMIE can outperform in all the metrics on the four datasets.
\end{abstract}


\end{frontmatter}


\section{Introduction}
\label{intro}
Topic models aim at discovering latent \emph{semantic topics} from a corpus of text documents and have been widely employed in information retrieval and related fields. 
The field of topic modeling has shifted away from ``Bag-of-Words'' representations such as Latent Dirichlet Allocation (LDA) \cite{blei2003latent} to neural networks based methods~\cite{MiaoYB16,MiaoGB17,esmaeili2019structured,abs-2004-12331}, which achieve state-of-the-art performance. 



Srivastava and Sutton \cite{SrivastavaS17} use an Autocoder-based topic model which constructs a Laplace approximation to the Dirichlet prior, and the proposed ProdLDA uses product of experts to learn topic representations. Similarly, a number of VAE-based neural topic models have also been proposed~\cite{MiaoYB16,MiaoGB17,CardTS17,GuiLPZXH19}. Different from the aforementioned neural topic model based on Gaussian distributions, Esmaeili et al \cite{esmaeili2019structured} use a neural topic model (VALTA) with Gumbel-Softmax\cite{JangGP17} to simulate discrete distributions. The advantage of using Gumbel-Softmax is that it promotes sparsity and leads to more disentangled representations, i.e.\ topics \cite{Burkhardt019}. 
Another type of widely used neural topic models are based on Generative Adversarial Networks (GAN)~\cite{goodfellow2014generative}. Wang et al~\cite{WangZH19} propose the Adversarial neural Topic Model (ATM) that is based on adversarial training. Moreover, they also propose the Bidirectional Adversarial Topic (BAT) method~\cite{abs-2004-12331}  that models topics with the Dirichlet prior and builds a two-way transformation between the document-topic distribution and the document-word distribution via bidirectional adversarial training.
 
A main limitation of the existing neural topic models is that they only try to constrain the learned topic representations without regarding the useful information conveyed by the input documents. The useful information is the representative information to distinguish the text from others. For example, in the text which are composed of words, the topics of the text and its words can convey the useful representative information which is informative in text clustering and classification. Here, a \emph{good} representation is one that can retain as much useful information of the input text as possible~\cite{goodfellow2014generative}. In topic modeling, the amount of useful information in learned topic representations is important for the tasks, i.e., topic distribution learning and topic word mining. To alleviate this problem, a simple and effective way is to train a representation learning network to maximize the mutual information (MI)~\cite{viola1997alignment}, between the input documents and their latent topic representations. Since mutual information can characterize both the relevance and the redundancy between random variables, it can effectively model the association of different variables. However, mutual information is difficult to estimate, especially in high-dimensional and continuous settings. 

In this paper, we introduce deep mutual information estimation~\cite{BelghaziBROBHC18} to topic modeling. Our method, Neural Topic Modeling with Deep Mutual Information Estimation (NTM-DMIE), effectively estimates and maximizes mutual information between high-dimensional input (document) and output (topic) pairs with deep neural networks. 
To best utilize the rich information contained in documents in learning topic representations, we regard documents as \emph{global} information and words contained in documents as \emph{local} information. 
Globally, we maximize MI between the documents attached with negative examples and the learned topic representations. Locally, NTM-DMIE maximizes the average MI between topic representations with document words to further improve the representation quality. 

The main contributions of our work are summarized as follows:
\begin{itemize}
\item We propose a novel neural topic modeling technique with deep mutual information estimation to better utilize document information. To the best of our knowledge, this is the first work to incorporate deep mutual information estimation into topic learning to improve the quality of topic representations and topic mining tasks. 
\item We propose to use global and local mutual information maximization to preserve the rich information contained in documents for learning their latent topic representations.
\item Extensive experimental results on four benchmark datasets 
show that our NTM-DMIE model outperforms recent, strong baseline methods.
\end{itemize}

\section{Related Work}
\label{rwk}
Our work is mainly related to neural topic modeling and deep mutual information estimation. We briefly discuss their recent progress. 

\subsection{Neural Topic Modeling}
Recently, neural networks\cite{kingma2013auto,goodfellow2014generative} have been employed in topic modeling, which are more effective and efficient at approximating the hidden, complex variables in the topic models. 
Based on Variational Auto-encoder (VAE), Miao et al \cite{MiaoYB16} proposed the Neural Variational Document Model (NVDM), which builds a deep neural network conditioned on text to approximate the intractable distributions over the latent variables. 
Moreover, they \cite{MiaoGB17} further proposed the Gaussian Softmax topic model (GSM), parameterized with neural networks. 
NVDM was extended for generalizing topic models to model with covariates, interactions, and customized regularizers \cite{CardTS17}. 
Card et al \cite{SmithCT18} developed a supervised neural topic model (\textsc{scholar}) which models metadata as a covariate or a predicted variable. 
ProdLDA \cite{SrivastavaS17} and  Variational Aspect-based Latent Topic
Allocation (VALTA) \cite{esmaeili2019structured} also embed relationships between documents, topics, and words in differentiable functions. 
Moreover, Neural Topic Model (NTM) \cite{ding2018coherence} and  Variational Topic Model with Reinforcement Learning(VTMRL) \cite{GuiLPZXH19} both incorporated topic coherence into topic modeling. 

Some neural topic models were designed based on Generative Adversarial Network(GAN), 
Wang et al \cite{WangZH19,abs-2004-12331,abs-2004-12331} proposed the Adversarial neural Topic Model (ATM) based on adversarial training.
Gupta et al \cite{GuptaCBS19} proposed a neural autoregressive topic model, DocNADE, to exploit the full context information around words in a document in a language modeling fashion. 

With the rapid development of topic-aware related work, topic models were designed with many other machine learning methods. Zhao X et al \cite{zhao2021neural} proposed the Variational Auto-Encoder Topic Model (VAETM) by combining word vector representation and entity vector representation to address the limitations for mining high-quality topics from short texts. Panwar M et al \cite{panwar2020tan} proposed the Topic Attention Networks for Neural Topic Modeling(TAN-NTM), which processed document as a sequence of tokens through an LSTM whose contextual outputs are attended in a topic-aware manner. Bahrainian S A et al \cite{bahrainian2021self} proposed a new light-weight Self-Supervised Neural Topic Model (SNTM) that learns a rich context by learning a topic representation jointly from three co-occurring words and a document that the triplet originates from. Jin Y et al \cite{jin2021neural} proposed a variational autoencoder (VAE) NTM model that jointly reconstructs the sentence and document word counts using combinations of bag-of-words (BoW) topical embeddings and pre-trained semantic embeddings. Zhao H et al \cite{zhao2020neural} proposed to learn the topic distribution of a document by directly minimising its OT distance to the document’s word distributions. Ma Z et al \cite{ma2021semantic} proposed a novel topic model named Semantic-based Bidirectional Adversarial Neural Topic Model (SNTM), which introduces semantic information into Bidirectional Generative Adversarial Networks (BiGAN) by adding the word embedding and BiLSTM-Attention mechanism. Wang Y et al \cite{wang2021layer} developed a novel neural topic model, namely Layer-Assisted Neural Topic Model (LANTM), to enhance the topic represen- tation encoding by not only using text contents, but also the assisted network links. Yang Y et al \cite{yang2021topnet} proposed TopNet, to leverage the recent advances in neural topic modeling to obtain high-quality skeleton words to complement the short input. Gupta P et al \cite{gupta2021multi} proposed a neural topic modeling framework using multi-view embedding spaces: pretrained topic-embeddings, and pretrained word-embeddings (context-insensitive from Glove and context-sensitive from BERT models) jointly from one or many sources to improve topic quality and better deal with polysemy.

Despite the continual research of neural topic modeling, existing works do not yet fully exploit the useful information contained in documents. 
Our method is the first neural topic model that employs deep mutual information estimation \cite{HjelmFLGBTB19}. It differs from the aforementioned neural topic models in the following main ways. 
(1) Unlike GSM, NVDM, and ProdLDA with a Gaussian or a logistic prior, we approximate the discrete topic assignment from a continuous distribution with Gumbel-Softmax\cite{JangGP17}, which can approximate categorical samples and whose parameters can be easily computed via the reparameterization trick. 
(2) Taking advantage of deep mutual information's capability of modeling the non-linear statistical dependence of the documents and the latent topics, our model estimates and maximizes mutual information between documents and topics to learn better representations of documents and topics.

\subsection{Deep Mutual Information Estimation}

One core objective of topic modeling is to learn useful topic representations. Similarly, deep mutual information estimation also aims to train an encoder for representation learning to maximize the mutual information (MI) \cite{viola1997alignment} between its input and output. To the best of our knowledge, there is no work using Deep Mutual Information Estimation on topic modeling, hence we briefly survey recent research work on deep mutual information estimation. 
Belghazi et al \cite{BelghaziBROBHC18} proposed MINE, a method to compute mutual information with neural network. 
Hjelm et al \cite{HjelmFLGBTB19} proposed Deep InfoMax (DIM) to estimate and maximize the mutual information between input data, with global and local information, and its high-level representation with adversarial learning \cite{goodfellow2014generative}. 
Yang et al \cite{YangDZYL19} proposed a dual autoencoder network with mutual information estimation to learn the robust and discriminative latent representations for deep spectral clustering. 
Guo et al \cite{GuoHKH19} proposed a method to learn disentangled representations, which incorporates deep mutual information estimation into the objective of cross-modal retrieval. Sanchez et al \cite{abs-1912-03915} proposed a method to learn the disentangled representations of images via deep mutual information estimation. 
Bachman et al \cite{BachmanHB19} proposed a method, Augmented Multiscale Deep InfoMax (AMDIM), for representation learning based on maximizing mutual information between features from multiple views of a shared context and the latent representation. 
Qian et al \cite{QianC19}  proposed VAE-MINE, which incorporates mutual information estimation (MINE) into variational autoencoder (VAE), to learn the latent representation. 
Zhou et al \cite{abs-2003-11521} proposed Text Matching with Deep Info Max (TIM), which maximizes the mutual information between the input and output with global and local information, for text matching. 
Benefiting from these work, we want to incorporate deep mutual information estimation into topic modeling to learn the topic representation more specific to given documents or sentences.

\section{Neural Topic Model with Deep Mutual Information Estimation}
\label{appro}
\begin{figure}[t]
\centering
\includegraphics[width=8.5cm,height=6cm]{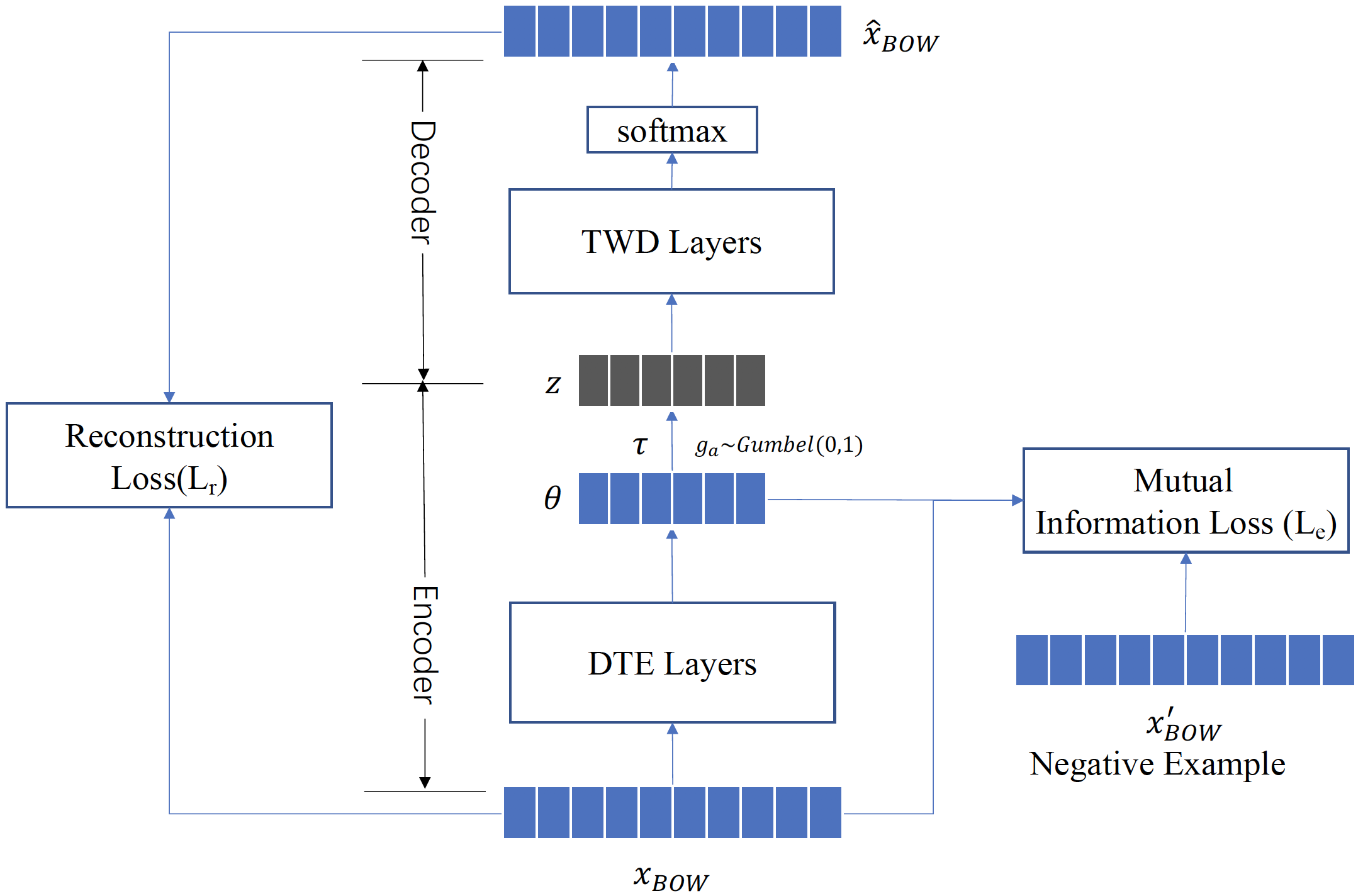}
\caption{The overall framework of  NTM-DMIE\label{fig:frm}}
\end{figure}

The overall framework of our Neural Topic Modeling with Deep Mutual Information Estimation (NTM-DMIE) can be seen in Figure \ref{fig:frm}. Our framework consists of two main components, i.e., Document-Topic Encoder and Topic-Word Decoder. (1) The Document-Topic Encoder, which learns robust latent topic representations of documents with the documents themselves and their negative examples, simulates the document-topic distribution in LDA. To preserve the rich information of documents, we introduce mutual information estimation between the documents and their topic representations in the encoder. (2) The Topic-Word Decoder, which embeds the latent topic representations into document words, learns the topic-word distribution as in LDA. These two components are jointly optimized in a unified framework. The generative process of the common topic modeling is given as follows: 

\begin{itemize}
    \item Draw a topic distribution $\theta \sim Dirichlet(\alpha)$ ,   
    \item For each word in the document, draw $w_n\sim Multinomial(\sigma(\beta\theta))$.
\end{itemize} 
Where $\alpha$ is the parameter of the Dirichlet distribution, $w_n$ is the $n$-th word in the document, $\beta$ is the topic word distribution, $\sigma$ is the softmax function, and $\theta$ and $\beta$ are the parameters of the document-topic and topic-word distributions respectively, which are computed with neural network in our work.

Let $x_{BOW}\in V^{\mathbb{Z}_{\geq0}}$ denote an input document in the bag-of-words representation, where $V$ is the vocabulary and $\mathbb{Z}_{\geq0}$ denotes non-negative integers. $x'_{BOW}$ denotes an negative example for learning the discriminative topic representation of $x_{BOW}$ and $\hat{x}_{BOW}$ is the document reconstructed by the autoencoder. $\hat{\theta}$ denotes the topic distribution of the documents and $\theta$ is the distribution after normalization. $z$ denotes their corresponding topic assignment based on the topic distribution $\theta$, and $K$ is the topic number and the dimension of $\theta$, $\hat{\theta}$ and $z$. Moreover, $\tau$, the temperature parameter of Gumbel-Softmax, and $g_a$, the base distribution, are used for sampling Gumbel-Softmax in the reparameterization trick. We use $x$ and $x_{BOW}$ interchangeably when there is no ambiguity. 

Our goal is to train a document encoder with deep mutual estimation \cite{BelghaziBROBHC18} to learn robust, discriminative topic representations for a given document together with its negative examples.

\subsection{Document-Topic Encoder}\label{sec:dte}
Learning the topic representation of documents is the core part of neural topic modeling and a good topic representation can improve the quality of learning the topic word distribution. 
Mutual information, as shown in Eq (\ref{eq:mutual}), can model the essential correlation between two objects. We use it to measure the association between documents $x \in X$ and latent topics $z \in Z$ to learn robust topic representations, where $X = \{x_1, ..., x_n\}$ denote the input documents, $Z =\{z_1, ..., z_n\}$ denote their corresponding latent topic representations.
\begin{equation}
\label{eq:mutual}
\begin{aligned}
I(X,Z)&=\int\int p(z|x)p(x)\log\frac{p(z|x)}{p(z)}dxdz\\
      &=KL(p(z|x)p(x)||p(z)p(x))
\end{aligned}
\end{equation}

In Eq \ref{eq:mutual}, $p(x)$ is the distribution of documents, $p(z|x)$ is the distribution of the latent topic and the marginal distribution of topic $p(z)$ is computed by $p(z)=\int p(z|x)p(x) dx$. The objective of our encoder is to maximize the mutual information as Eq (\ref{eq:max}). To make the learned topic {more representative to the documents}, the marginal distribution of latent topic $p(z)$ must obey the prior distribution of Dirichlet distribution $q(z)$. 
\begin{equation} 
\label{eq:max}
p(z|x)=\max \limits_{W_E}\{I(X,Z)\}
\end{equation}
where $W_E$ is the parameter of the encoder $E$. 


Based on Eq (\ref{eq:max}), the objective of the encoder can be summarized as Eq (\ref{eq:obe}). 

\begin{equation} 
\label{eq:obe}
\begin{aligned}
\hat{p}(z|x)=\min \limits_{W_E}\{&-\beta I(X,Z)\\
&+\gamma \int\int p(z|x)p(x)\log \frac{p(z|x)}{q(z)}dxdz\}
\end{aligned}
\end{equation}



The first term of Eq (\ref{eq:obe}) is the mutual information defined in Eq (\ref{eq:mutual}), and the second term is the KL divergence of the posterior $p(z|x)$ and the prior $q(z)$, which is beneficial to make the latent topic space more regular. Here, $\gamma$ is the smoothing parameter of mutual information and KL divergence. 
To resolve the problem of unbounded KL divergence in the calculation of mutual information (Eq (\ref{eq:mutual})), the Jensen-Shannon (JS) divergence is employed for mutual information estimation. The objective thus can be rewritten as Eq (\ref{eq:js}).

\begin{equation} 
\label{eq:js}
\begin{aligned}
\hat{p}(z|x)=\min \limits_{W_E}\{&-\beta JS(p(z|x)p(x),p(z)p(x))\\
&+\gamma E_{x\sim p(x)}[KL(p(z|x)||q(z))]\}
\end{aligned}
\end{equation}

Inspired by Nowozin et al \cite{nowozin2016f}, we also use the variational estimation of JS divergence with adversarial learning. The first term of Eq (\ref{eq:js}) can thus be optimized as Eq (\ref{eq:jst}). 

\begin{equation} 
\label{eq:jst}
\begin{aligned}
\max \limits_{T}\{&\mathbb{E}_{x,z\sim p(z|x)p(x)}[\log(\sigma(T(x,z)))]\\
+&\mathbb{E}_{x,z\sim p(z)p(x)}[\log(1-\sigma(T(x,z)))]\}
\end{aligned}
\end{equation}
where $T(x,z)$ is given in Eq (\ref{eq:jstt}) and $\sigma(T(x,z))$ is a discriminator.

\begin{equation} 
\label{eq:jstt}
T(x,z)=\log\frac{2p(z|x)p(x)}{p(z|x)p(x)+p(z)p(x)}
\end{equation}

\begin{figure}[t]
\centering
\includegraphics[width=7.5cm]{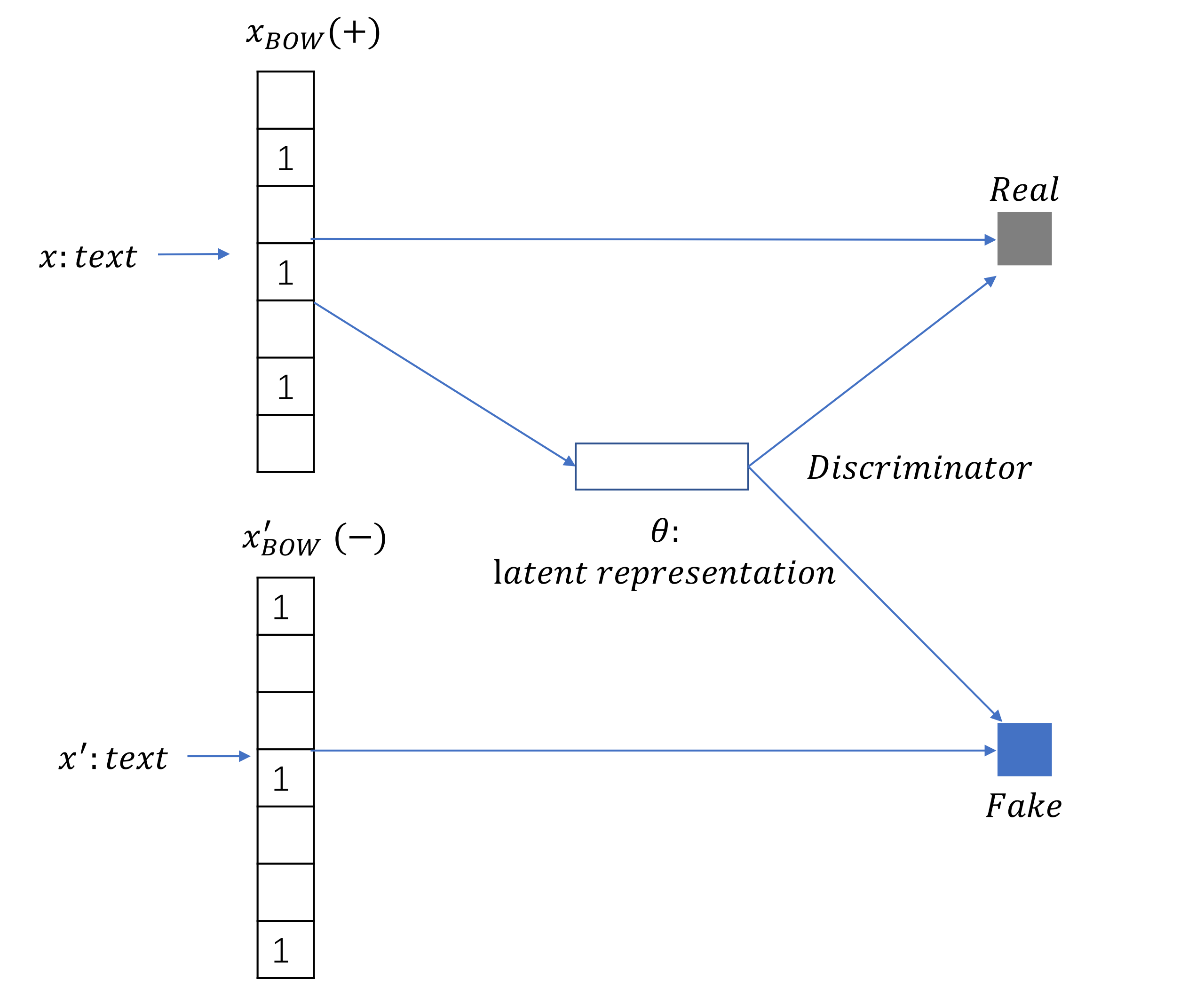}
\caption{Global information: document and its negative examples. The latent topic representation is concatenated with the global “Bag of Words” representation. A 1 × 1 convolutional discriminator is used to score the ‘real’ document and its topic representation, while ‘fake’ is the randomly selected document with the learnt topic representation.\label{fig:global}}
\end{figure}

\begin{figure}[t]
\centering
\includegraphics[width=8cm]{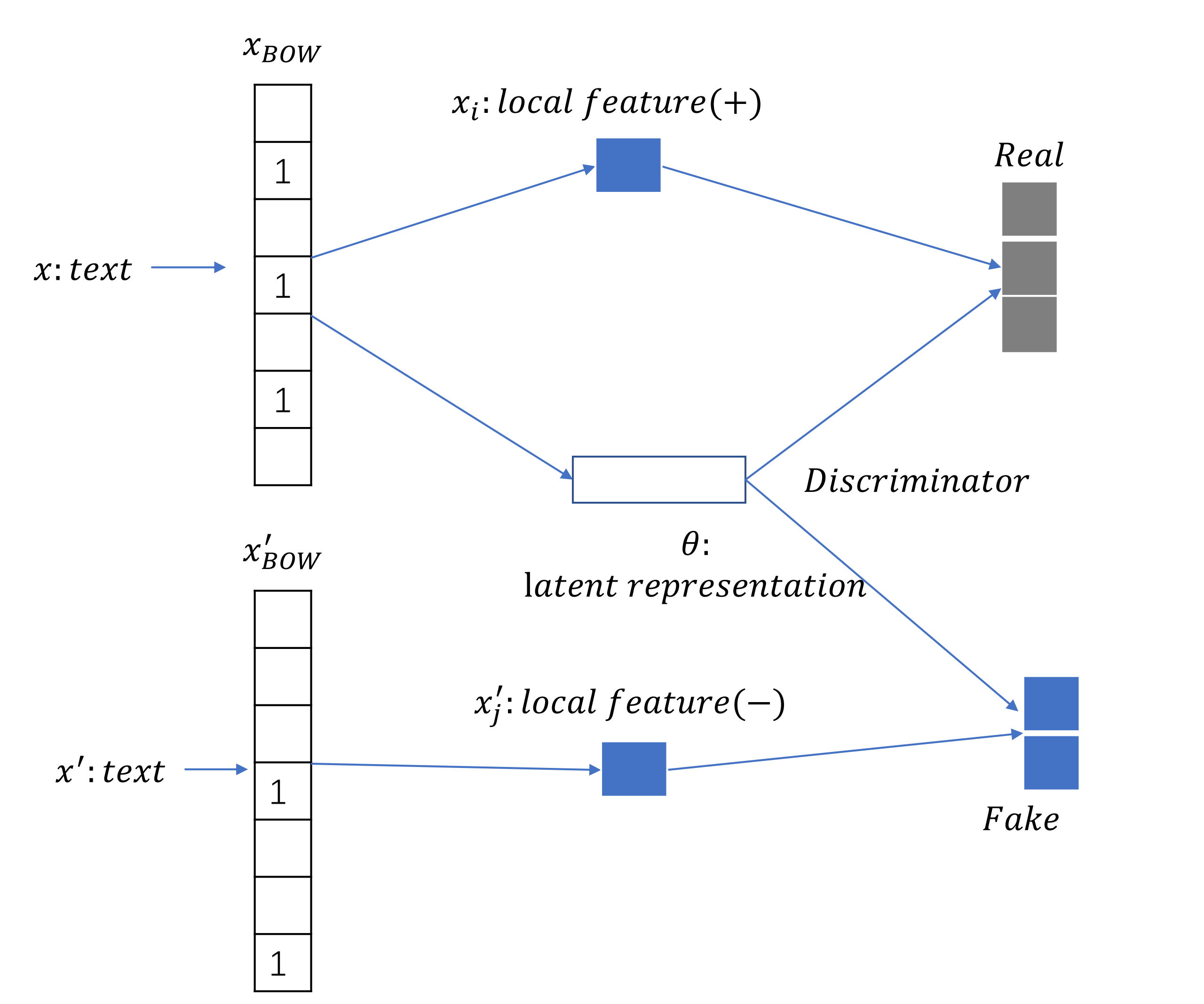}
\caption{Local information: document word and its negative examples. The latent topic representation is concatenated with the local word information. ‘real’ corresponds to the pair with word in the given document and the topic representation of the document, while ‘fake’ corresponds to the the pair with the randomly selected word and the given topic representation.\label{fig:local}}
\end{figure}

To learn robust topic representations, the encoder uses a discriminator to differentiate the original document and its negative examples to measure the topic representation of the original document. In Eq (\ref{eq:jst}), $\sigma(T(x,z))$ is used for discriminating the original document $x$ and its negative examples $\{x'\}$, where the negative examples are selected from the corpus with a deterministic strategy (which we will describe below). 
Eq (\ref{eq:jst}) only considers the global mutual information: that between the whole document $x$ and its topic representation $z$, as shown in Figure \ref{fig:global}. The words in the document, i.e.\ local information, also play an important role in learning the topic representation of the document. Hence, we introduce the local mutual information to model the association between document words and the topic representation as shown in Figure \ref{fig:local}. Similar to the global mutual information, negative examples are also used in the local mutual information. The loss function of the encoder is given as Eq (\ref{eq:le}), where $|x|$ is the number of words in the document, $x_i$ represents the $i$-th word in the document and $q(z)$ is the symmetrical Dirichlet distribution as the prior distribution.

\begin{equation} 
\label{eq:le}
\begin{aligned}
L_e=-\beta(&\mathbb{E}_{x,z\sim p(z|x)p(x)}[\log(\sigma(T(x,z)))]
\\+&\mathbb{E}_{x,z\sim p(z)p(x)}[\log(1-\sigma(T(x,z)))])
\\-&\frac{\beta}{|x|}\sum_{i}(\mathbb{E}_{x,z\sim p(z|x)p(x)}[\log(\sigma(T(x_i,z)))]
\\+&\mathbb{E}_{x,z\sim p(z)p(x)}[\log(1-\sigma(T(x_i,z)))])
\\+&\gamma \mathbb{E}_{x\sim p(x)}[KL(p(z|x)||q(z))]
\end{aligned}
\end{equation}

In Eq (\ref{eq:le}), $q(z)$ is the standard prior Dirichlet distribution and $p(x)$ is given by the corpus. The core part, $p(z|x)$, is the encoder of Fig \ref{fig:frm}, and $p(z)$ can be computed via $\int p(z|x)p(x)dx$. To model $p(z|x)$, note that the input $x$ is represented as $x_{BOW}$. Then, a feedforward neural network with two hidden layers (FC Layers) is utilized to embed $x_{BOW}$ into a hidden vector $\hat{\theta}$. To solve the problem of \emph{posterior collapse} in VAE \cite{abs-2004-12585}, a batch normalization layer \cite{Burkhardt019} is added and the hidden vector $\hat{\theta}$ with normalization is transformed as $\theta$, which is the latent representation of the document $x$, and the parameter of the discrete distribution, i.e.\ $p(z|x)$. In the case of the Dirichlet distribution, we use the concrete distribution, a relaxation of discrete distribution via a Gumbel-Softmax \cite{esmaeili2019structured}, for sampling via the reparameterization trick. 


\subsection{Topic-Word Decoder}
The decoder of NTM-DMIE, the TWD layers of Fig \ref{fig:frm}, is a linear transformation layer that maps $z$ for document $x$ to the predicated probability of words $\hat{x}$, i.e., the reconstructed document. The reconstruction loss mainly depends on two parts: the distribution of latent topics and the generative performance of the decoder network. Our goal is to compute the word distribution of each topic $\tau$ via the decoder network. The reconstructed document can be obtained by Eq (\ref{eq:re}), where $\tau\in \mathbb{R}^{|V|\times K}$. Each column of $\tau$, $\tau_k$, represents the word distribution of the topic $k$.

\begin{equation}
\label{eq:re}
p(\hat{x}|\tau,z)=softmax(\tau z+b)
\end{equation}

The reconstruction loss is defined over the reconstructed document $\hat{x}$ and the original document $x$ as below.
\begin{equation}
\label{eq:loss}
L_r=||\hat{x}-x||_F^2
\end{equation}

\subsection{Model Training}
The entire NTM-DMIE model is trained in an end-to-end manner. 
The overall loss function $L$ is a weighted sum of the mutual information loss and the reconstruction loss. 
\begin{equation}
\label{eq:tloss}
L=\mu*L_r+(1-\mu)*L_e
\end{equation}

\section{Experimental Setup}
\label{sect:es}
\subsection{Dataset}
\label{sect:ds}
We evaluate the performance of NTM-DMIE on four public datasets, including two labeled datasets and two unlabeled datasets. 
\textbf{20 Newsgroups\footnote{\url{http://qwone.com/~jason/20Newsgroups/}}.} is a collection of approximately 20,000 newsgroup documents, partitioned (nearly) evenly across 20 different newsgroups. Documents in the 20 newsgroups collections have class labels, so the dataset is used for evaluating the performance of text classification. \textbf{AG\_News} is a collection of more than one million news articles which include four different groups: World, Sports, Business and Sci/Tec. News articles have been gathered from more than 2,000 news sources by ComeToMyHead in more than 1 year of activity. We selected its subset, which contains 12,760 articles, for the evaluation. \textbf{NY times} is a collection of news articles published between 1987 and 2007, and contains a wide range of topics, such as sports, politics, education, etc. \textbf{Wikitext-103} is a collection of more than 100 million sentences, all extracted from Wikipedia's Good and Featured articles. It has been widely used in language modeling. \begin{table}[h]
\centering
\caption{Statistics of the four datasets.}
\label{tab:dataset}
\resizebox{.6\textwidth}{!}
{
\begin{tabular}{lccc}
\toprule
Dataset&Num of Docs&Average\_Size&Num of Labels\\ 
\midrule
20NG&19,999&220.5&20\\
AG\_News&12,760&26.3&4\\
NYTimes&242,798&7.29&-\\
Wikitext-103&307,807&176.2&-\\
\bottomrule
\end{tabular}
}
\end{table}

We conducted the following common preprocessing steps: conversion into lowercase, word tokenization, lemmatization, and removal of stop words special characters. 
After preprocessing, the statistics of the four datasets are summarized in Table~\ref{tab:dataset}.

\subsection{Baselines}
\label{sect:bs}
We compared our NTM-DMIE model with the following state-of-the-art methods:
\begin{itemize}
    \item \textbf{ProdLDA}~ \cite{SrivastavaS17} is an Autocoder-based topic model that constructs a Laplace approximation to the Dirichlet prior.
    \item \textbf{GSM}~\cite{MiaoGB17} is a Gaussian Softmax topic model parameterized with neural networks.
    \item \textbf{NTM}~\cite{ding2018coherence} is a neural topic model which incorporates a topic coherence objective. 
    \item \textbf{Scholar}~\cite{SmithCT18} is a supervised neural topic model that allows for metadata to appear as either a covariate or a predicted variable in the model structure.
    \item \textbf{Gaussian-BAT}~\cite{abs-2004-12331} models topics with the Dirichlet prior and builds a two-way transformation between document-topic distribution and document-word distribution via bidirectional adversarial training. 
\end{itemize}

\subsection{Implementation Details}
The hyperparameters of ProdLDA, GSM, NTM, Scholar and Gaussian-BAT are set according to the best hyperparameters reported in their original papers. Topic number was set as $10$, $20$ and $50$ for all the four datasets to test the ability of text clustering and the quality of topics in our model compared with the baselines. During model training, we used the Adam optimizer with a learning rate of  $10E-4$ on the all datasets. For the weight of the loss function of the encoder given in Eq \ref{eq:le}, it was set as $\beta=\gamma=1$. For $\mu$ of the overall loss function in Eq (10), it was empirically set as $\mu = 0.4$. All the experiments were conducted for 100 epochs with batch size 128. And all models are implemented by PyTorch with a single Nvidia GTX 1080Ti graphic card, running for four times. In our model, negative examples are used for learning robust topic representations of the document. In our experiment, we use different strategies for selecting negative examples: random selection and similarity-based selection. Hence, our model have two variants, NTM-DMIE (random) and NTM-DMIE (similarity). 

\subsection{Evaluation Metrics}
\label{sect:em}
We compare model performance on \emph{topic coherence} and \emph{topic uniqueness} to evaluate the quality of topics. We also perform text clustering to measure the reconstruction ability of latent features, for which we use accuracy. 

\textbf{Topic Coherence (NPMI)}~\cite{aletras2013evaluating} Topic coherence indicates that the words in a topic should be as coherent as possible. For this we use the widely-used metric Normalized Pointwise Mutual Information (NPMI), which assumes coherent words should co-occur within a certain distance. Given the top $M$ topic words ordered by their probabilities, the NPMI score of the topic can be calculated as follows:
\begin{equation}
    NPMI = \frac{1}{M}\sum_{w_{i},w_{j}}^{}\frac{\log\frac{p(w_{i},w_{j})+\epsilon}{p(w_{i})p(w_{j})}}{-\log(p(w_{i},w_{j})+\epsilon)}
\end{equation}
where $p(w_{i})$ is the probability of word $w_{i}$, $p(w_{i},w_{j})$ is the co-occurrence probability of $w_{i}$, $w_{j}$ within a window in the reference corpus and $\epsilon$ is used to avoid division by zero.

\textbf{Topic Uniqueness (TU)}~\cite{nan-etal-2019-topic} measures the diversity of a set of topics, and can be used to determine how distinguished the topics are from each other. Given the top $M$ words for each of the $K$ topics, TU for topic $k=1,\ldots,K$ can be defined as follows:
\begin{equation}
\label{eq:tu}
TU(k) = \frac{1}{M}\sum_{i=1}^{M}\frac{1}{cnt(i,k)}
\end{equation}
where $cnt(i,k)$ is the occurrence count of the $i$-th top word in topic $k$ in the top words across all topics. The range of TU value is between $1/K$ and $1$. A higher TU value indicates that fewer words are repeated across topics, thus the produced topics are more diverse. 

\textbf{Clustering Accuracy (ACC)}~\cite{abs-2004-12331} measures the effectiveness of learned topics on document clustering, in which the learned topic distributions are used as features for clustering. 
Model performance is evaluated by accuracy (ACC) as follows:
\begin{equation}
\label{eq:acc}
ACC = \max_{mapping} \frac{\sum_{i=1}^{N_{t}}\mathbbm{1}(l_{i} = mapping(c_{i}))}{N_{t}}    
\end{equation}
where $N_t$ is the number of documents in the test set, $\mathbbm{1}()$ is the indicator function, $l_i$ is the ground-truth label of the $i$-th document, $c_i$ is the cluster assignment of the $i$-th document, and $mapping$ ranges over all possible one-to-one mappings between labels and clusters. A higher ACC score means that the model is more likely to capture features that are representative of the given corpus.

\section{Results and Analysis}
\label{sect:ra}

In this section we present a comprehensive empirical evaluation on our proposed method. Our experiment evaluate the metrics, topic coherence for evaluating the quality of the learnt topics, topic uniqueness for the diversity of the topics, and the text clustering for evaluating the representative ability of the topic representations. Finally, the qualitative analysis of the learnt topics are given below. The source code of our experiment is given in the link \href{https://github.com/Asuper-code/NTM-DMIE}{https://github.com/Asuper-code/NTM-DMIE}.

\subsection{Topic Coherence and Topic Uniqueness}
\label{ssec:tc}
\begin{figure}[t]
\centering
\subfigure[NPMI results.]{
\begin{minipage}[t]{0.5\linewidth}
\centering
\includegraphics[width=2in]{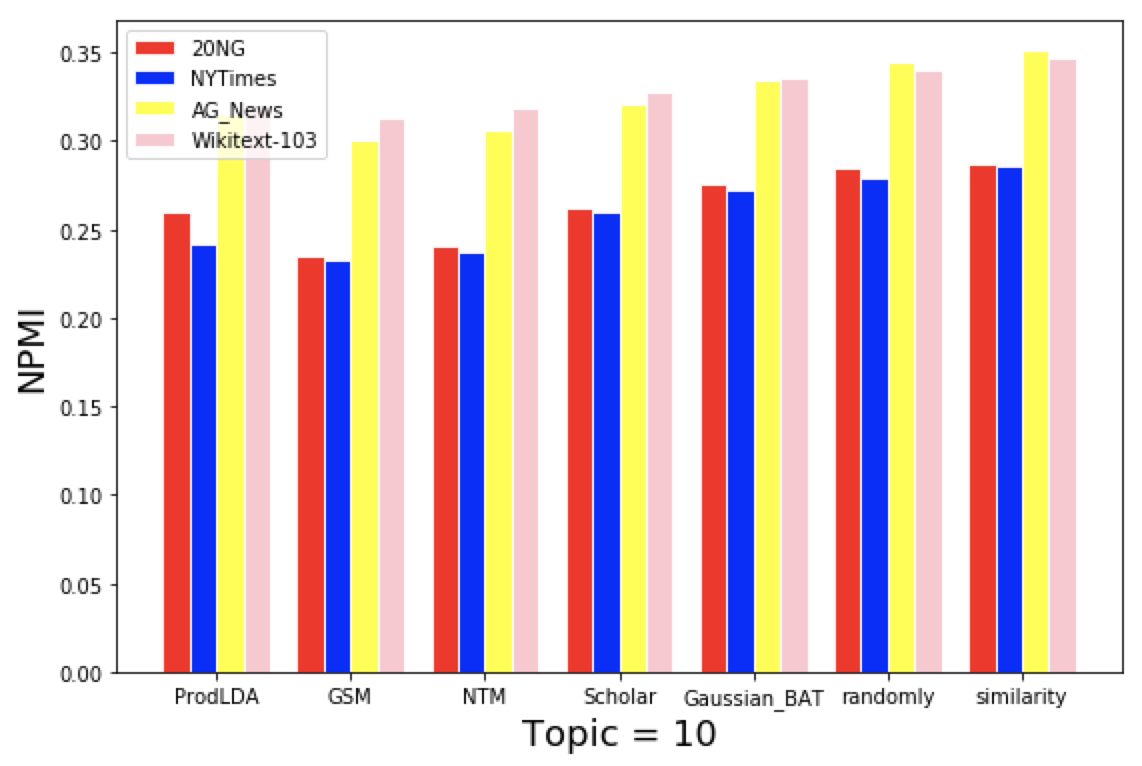}
\end{minipage}%
}%
\subfigure[TU results.]{
\begin{minipage}[t]{0.5\linewidth}
\centering
\includegraphics[width=2in]{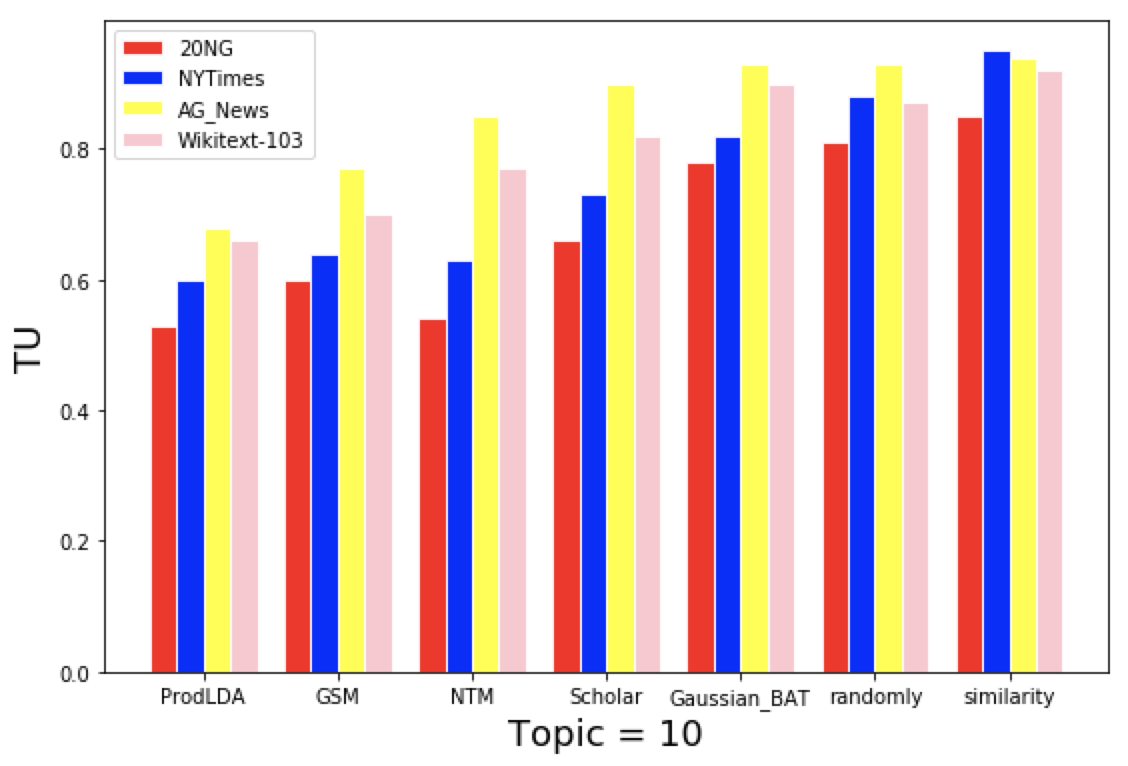}
\end{minipage}%
}%
\caption{ NPMI and TU Performance when Topic=10 on the four datasets.\label{fig:bar10} }
\end{figure}

\begin{figure}[t]
\centering
\subfigure[NPMI results.]{
\begin{minipage}[t]{0.5\linewidth}
\centering
\includegraphics[width=2in]{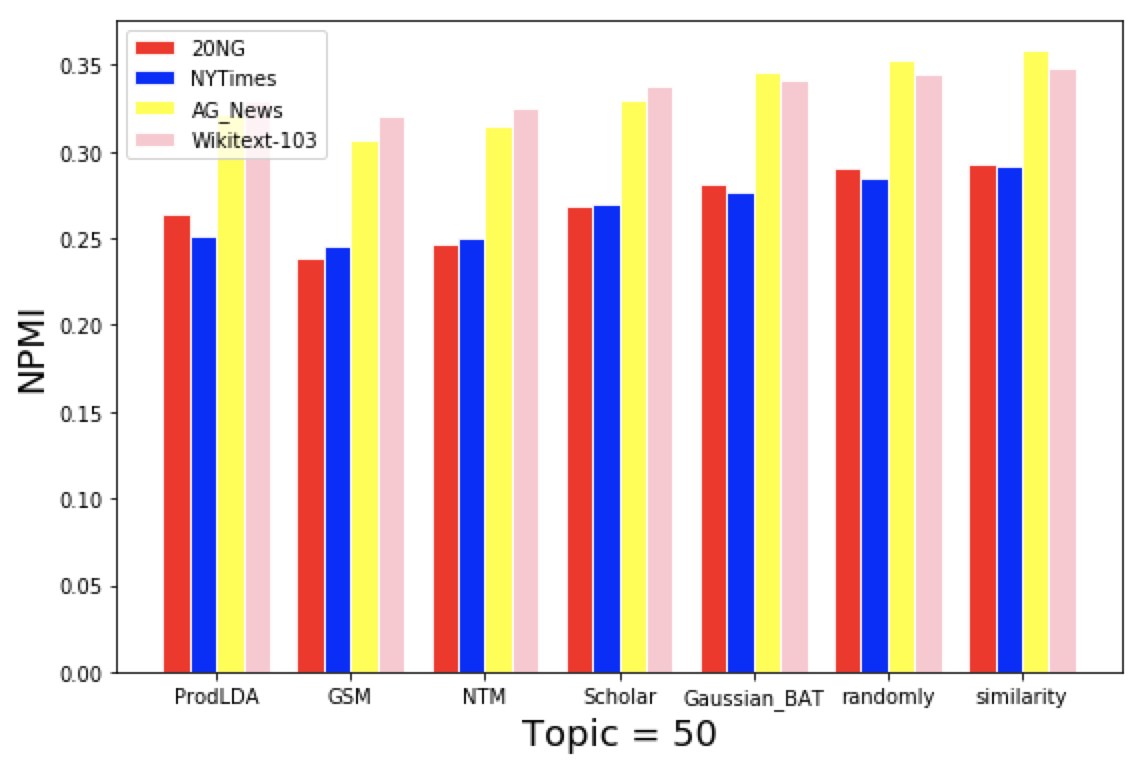}
\end{minipage}
}%
\subfigure[TU results.]{
\begin{minipage}[t]{0.5\linewidth}
\centering
\includegraphics[width=2in]{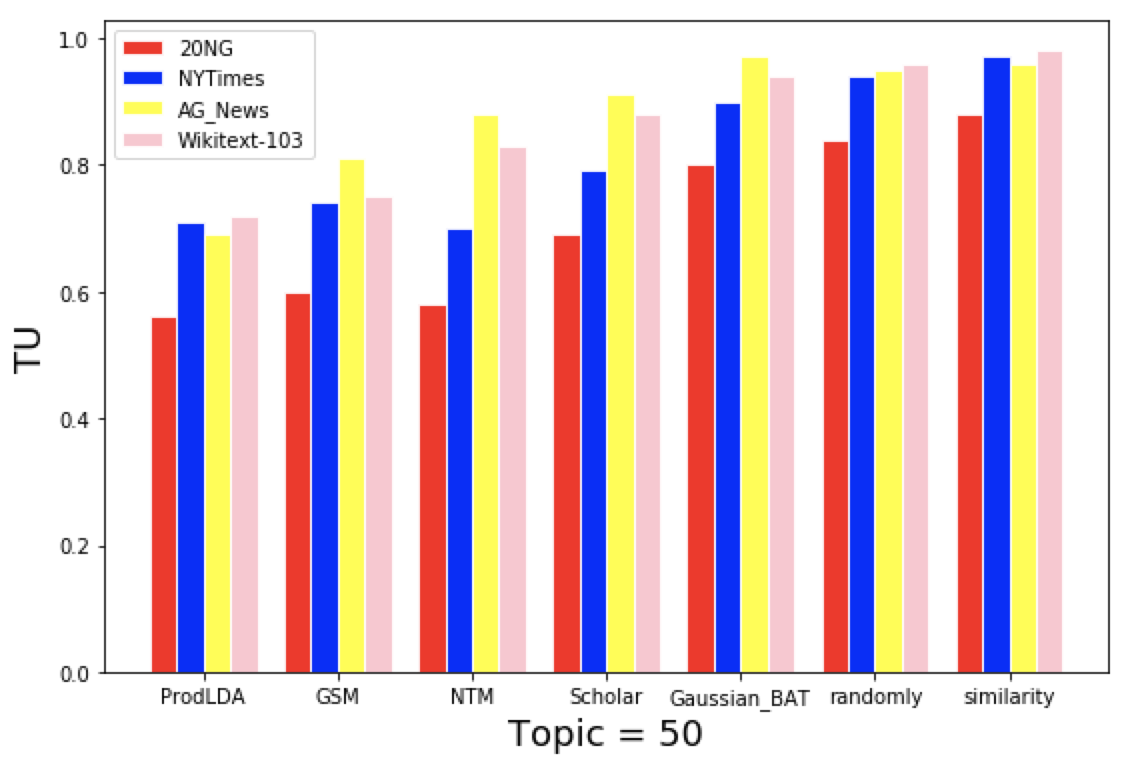}
\end{minipage}
}
\centering
\caption{ NPMI and TU Performance when Topic=50 on the four datasets.\label{fig:bar50} }
\end{figure}

\begin{figure}[t]
\centering
\subfigure[NPMI results.]{
\begin{minipage}[t]{0.5\linewidth}
\centering
\includegraphics[width=2in]{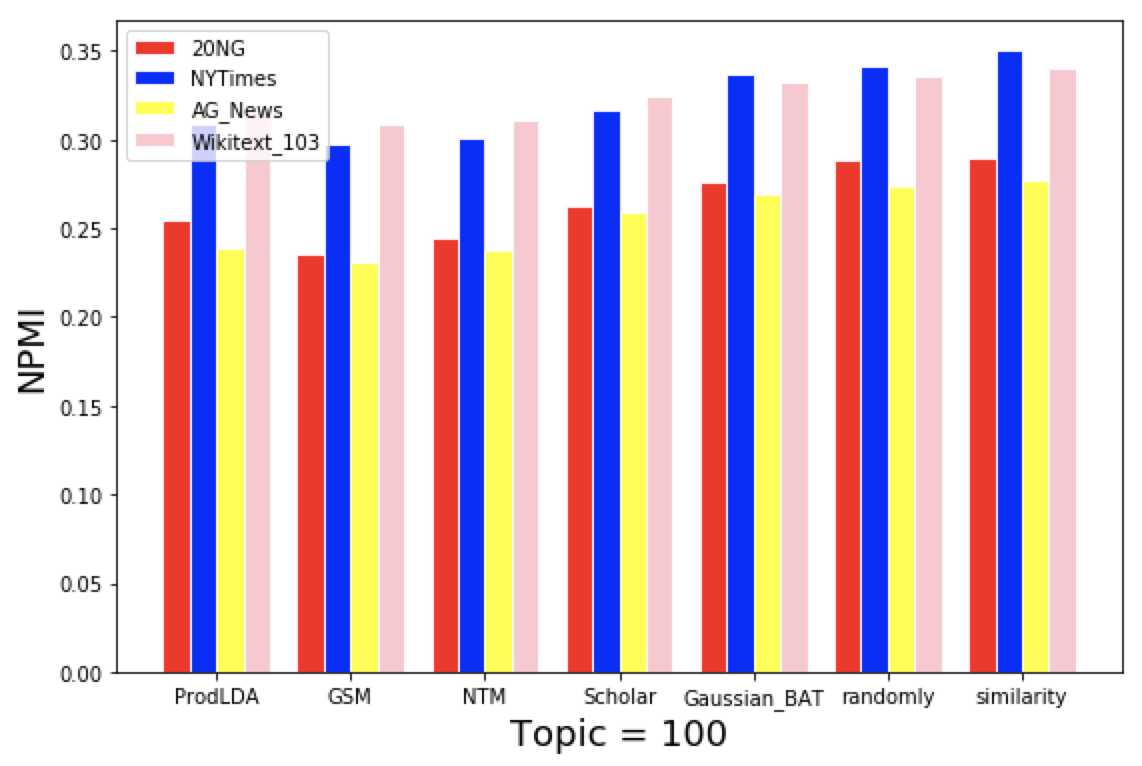}
\end{minipage}
}%
\subfigure[TU results.]{
\begin{minipage}[t]{0.5\linewidth}
\centering
\includegraphics[width=2in]{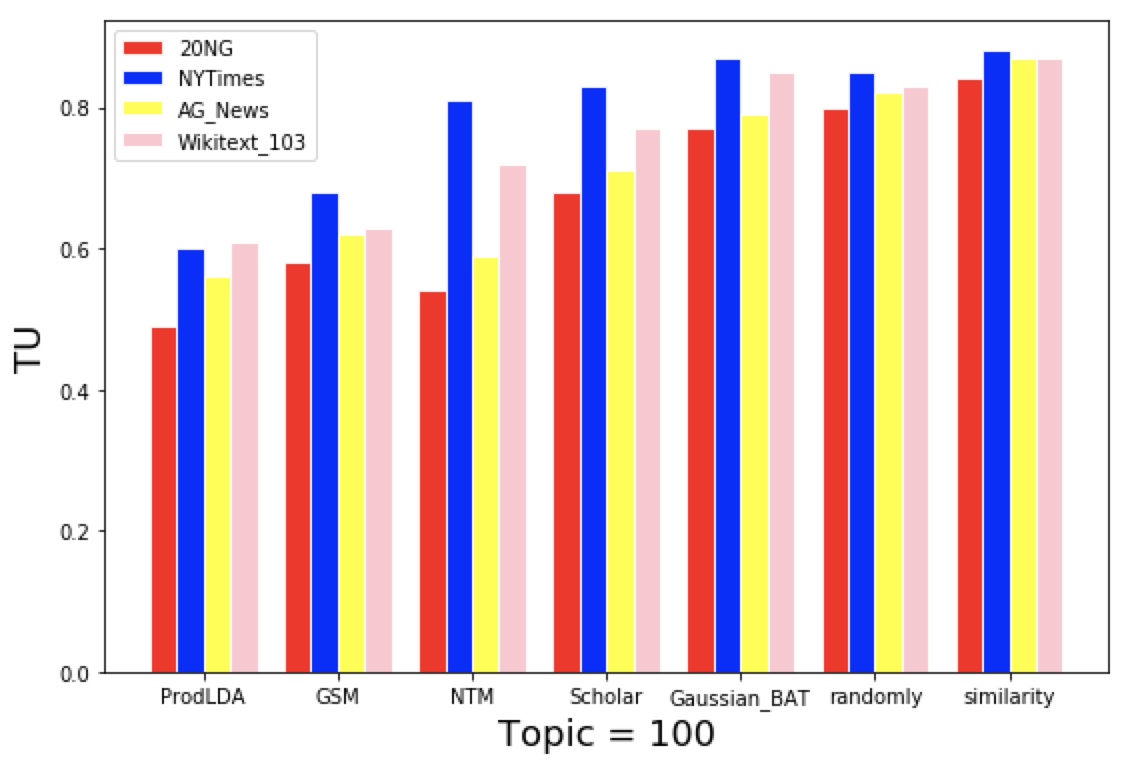}
\end{minipage}
}
\centering
\caption{ NPMI and TU Performance when Topic=100 on the four datasets.\label{fig:bar100} }
\end{figure}

Table~\ref{tab:npmitu} presents the results on topic coherence (measured by NPMI) and topic uniqueness with the number of topics set to 20.
Results with the number of topics set to $10$ , $50$ and $100$ can be found in Figure \ref{fig:bar10}, Figure \ref{fig:bar50} and Figure \ref{fig:bar100}, where our model exhibits similar superiority over the compared state-of-the-art models. Among Figure \ref{fig:bar10} , \ref{fig:bar50} and \ref{fig:bar100}, we also give the result of NMPI and TU of the two variants, NTM-DMIE(random) and NTM-DMIE(similarity). And the result show that NTM-DMIE(similarity) performed better than NTM-DMIE(random). The detailed analysis of the experiment about selection of negative examples will be discussed in section \ref{ssec:cs}.
\begin{table}[H]
    \centering
    \caption{Topic quality evaluation for 20 topics. The numbers in each cell are NPMI/TU, showing the 95\% confidence interval. Both NPMI and TU values are the higher the better.}
    \label{tab:npmitu}
    \resizebox{\textwidth}{!}
    {
    \begin{tabular}{lcccccccc}
    \toprule
   \multirow{2}{*}{Method}
    & \multicolumn{2}{c}{20NG} 
    &\multicolumn{2}{c}{NYTimes}
    &\multicolumn{2}{c}{AG\_News}
    &\multicolumn{2}{c}{Wikitext-103}\\
    \cmidrule(lr){2-3}\cmidrule(lr){4-5}\cmidrule(lr){6-7}\cmidrule(lr){8-9}
    {}&NPMI&TU&NPMI&TU&NPMI&TU&NPMI&TU\\
    \midrule
    ProdLDA &0.267$\pm$0.002&0.58$\pm$0.01&0.319$\pm$0.001&0.67$\pm$0.03&0.247$\pm$0.003&0.65$\pm$0.02&0.325$\pm$0.001&0.69$\pm$0.02\\
    GSM&0.243$\pm$0.001&0.65$\pm$0.02&0.303$\pm$0.002&0.79$\pm$0.02&0.239$\pm$0.003&0.70$\pm$0.03&0.317$\pm$0.001&0.71$\pm$0.02\\
    NTM&0.252$\pm$0.003&0.62$\pm$0.02&0.310$\pm$0.003&0.88$\pm$0.01&0.245$\pm$0.001&0.67$\pm$0.01&0.320$\pm$0.003&0.79$\pm$0.02\\
    Scholar&0.273$\pm$0.004&0.73$\pm$0.01&0.328$\pm$0.001&0.89$\pm$0.01&0.265$\pm$0.002&0.78$\pm$0.02&0.333$\pm$0.002&0.85$\pm$0.01\\
    Gaussian-BAT&0.285$\pm$0.001&0.85$\pm$0.01&0.344$\pm$0.002&0.95$\pm$0.01&0.274$\pm$0.001&0.86$\pm$0.01&0.338$\pm$0.003&0.93$\pm$0.02\\
    \midrule
    NTM-DMIE(random)&0.294$\pm$0.001&0.88$\pm$0.03&0.350$\pm$0.002&0.94$\pm$0.01&0.281$\pm$0.004&0.90$\pm$0.02&0.341$\pm$0.003&0.91$\pm$0.01\\
    NTM-DMIE(similarity)&\textbf{0.298}$\pm$0.002&\textbf{0.93}$\pm$0.01&\textbf{0.357}$\pm$0.003&\textbf{0.96}$\pm$0.01&\textbf{0.285}$\pm$0.002&\textbf{0.94}$\pm$0.02&\textbf{0.347}$\pm$0.001&\textbf{0.94}$\pm$0.02\\
    \bottomrule
    \end{tabular}
}
\end{table}

In Table \ref{tab:npmitu}, in terms of topic coherence, NTM-DMIE achieves the highest NPMI scores on all four datasets. Moreover, the NPMI of our model is substantially higher than all the baselines. This result demonstrates that our model can obtain more coherent topic words than the state-of-the-art baselines. 

As for topic uniqueness, NTM-DMIE also achieves the highest scores than all baselines on all the four datasets. 
This result indicates that our model can obtain topics with less repetition better than the baseline models.

\subsection{Ablation Study}
\label{ssec:as}
We conduct an ablation study to examine the effectiveness of the local information and global information components in our framework. Experiments are conducted on all four datasets with the topic number set to 20. 

In Table~\ref{tab:ablabel} and Table \ref{tab:abunlabel} (for labeled and unlabeled datasets respectively), the full NTM-DMIE model outperforms both variants significantly. It attests to the effectiveness of both local and global mutual information in our framework. On NPMI, we can see that NTM-DMIE with local information only performs better than with global information only, which can be attributed to the fact that local information helps the model capture more specific and high-quality features than global information. A similar observation can be made for TU, where local information performs better except a slightly worse TU score on the NYTimes dataset. 

In the Table~\ref{tab:ablabel} and Table \ref{tab:abunlabel}, it show the result of the variants of our model, which are the one without the global information,  the one without local information, and the one without both the global information and local information. Based on the result, we can find that the NPMI and TU performance of the model obviously decline when our model remove the global information or the local information.
\begin{table}[H]
    \centering
    \small
    \caption{Ablation study of labeled datasets.}
    \label{tab:ablabel}
    \resizebox{.7\textwidth}{!}{
    \begin{tabular}{lcccc}
    \toprule
    \multirow{2}{*}{Method}
    &  \multicolumn{2}{c}{20NG} &  \multicolumn{2}{c}{AG\_News}\\
    \cmidrule(lr){2-3}\cmidrule(lr){4-5}
    {}&NPMI&TU&NPMI&TU\\
    \midrule
    NTM-DMIE
    &\textbf{0.298}&\textbf{0.93}&\textbf{0.285}&\textbf{0.94}\\
    \quad w/ gloal information only
    &0.243&0.79&0.241&0.76\\
    \quad w/ local information only &0.252&0.82&0.247&0.79\\
    \quad w/ both 
    &0.223&0.66&0.225&0.61\\
    \bottomrule
    \end{tabular}
    }
    
\end{table}
\begin{table}[H]
    \centering
    \small
    \caption{Ablation study of unlabeled datasets.}
    \label{tab:abunlabel}
    \resizebox{.7\textwidth}{!}{
    \begin{tabular}{lcccc}
    \toprule
    \multirow{2}{*}{Method}
    &  \multicolumn{2}{c}{NYTimes} &  \multicolumn{2}{c}{Wikitext-103}\\
    \cmidrule(lr){2-3}\cmidrule(lr){4-5}
    {}&NPMI&TU&NPMI&TU\\
    \midrule
    NTM-DMIE
    &\textbf{0.357}&\textbf{0.96}&\textbf{0.347}&\textbf{0.94}\\
    \quad w/ gloal information only
    &0.319&0.88&0.323&0.80\\
    \quad w/ local information only &0.321&0.89&0.329&0.84\\
    \quad w/ both
    &0.300&0.71&0.307&0.63\\
    \bottomrule
    \end{tabular}
    }
\end{table}

\subsection{Text Clustering Performance}
\label{ssec:cp}
We evaluate the text clustering performance of the topic models learned by each method on the labeled datasets 20NG and AG\_News, and Table~\ref{tab:cluster} presents the overall results for all the models where the topic number is set to $20$, which is set based on the number of classes in the labeled datasets. 

In Table \ref{tab:cluster},  features captured by NTM-DMIE obtain the highest accuracy on text clustering on both datasets. We attribute this result to the mutual information framework of NTM-DMIE, which helps reconstruct the input and improves the quality of topics captured by the model. 
\begin{table}[h]
    \centering
    \caption{Text clustering performance of different methods on the 20Newsgroups dataset, where topic number is set to be $20$. Higher value indicates better performance.}
    \label{tab:cluster}
    \resizebox{.4\textwidth}{!}{
    \begin{tabular}{lcc}
    \toprule
    Model& 20NG&AG\_News \\[3pt]
    \midrule
    ProdLDA& 32.5\% &78.2\%\\[3pt]
    GSM& 31.7\% &80.1\%\\[3pt]
    NTM& 33.9\% &82.3\%\\[3pt]
    Scholar& 35.7\% &82.9\%\\[3pt]
    Gaussian\_BAT& 39.9\% &84.5\%\\[3pt]
    \midrule
    NTM-DMIE (random)&43.5\%&85.1\%\\[3pt]
    NTM-DMIE (similarity)&\textbf{45.2\%}&\textbf{85.9\%}\\[3pt]
    \bottomrule
    \end{tabular}}
\end{table}

\subsection{Selection Strategies of Negative Examples}
\label{ssec:cs}
We employ a discriminator in the document-topic encoder to distinguish a document from its negative examples for learning robust topic representations of the document.  
Here we examine the effect of different strategies for selecting negative examples: random selection and similarity-based selection. 

Random selection, as the name suggests, randomly chooses several negative examples for a given example. In contrast, the similarity-based selection strategy chooses the most dissimilar documents as negative examples.
The results of NPMI and TU on all four datasets are shown in Table~\ref{tab:npmitu}, with the topic number set to 20.

\begin{figure}[H]
\centering
\subfigure[Topic = 10\_NPMI.]{
\begin{minipage}[t]{0.45\linewidth}
\centering
\includegraphics[width=2in]{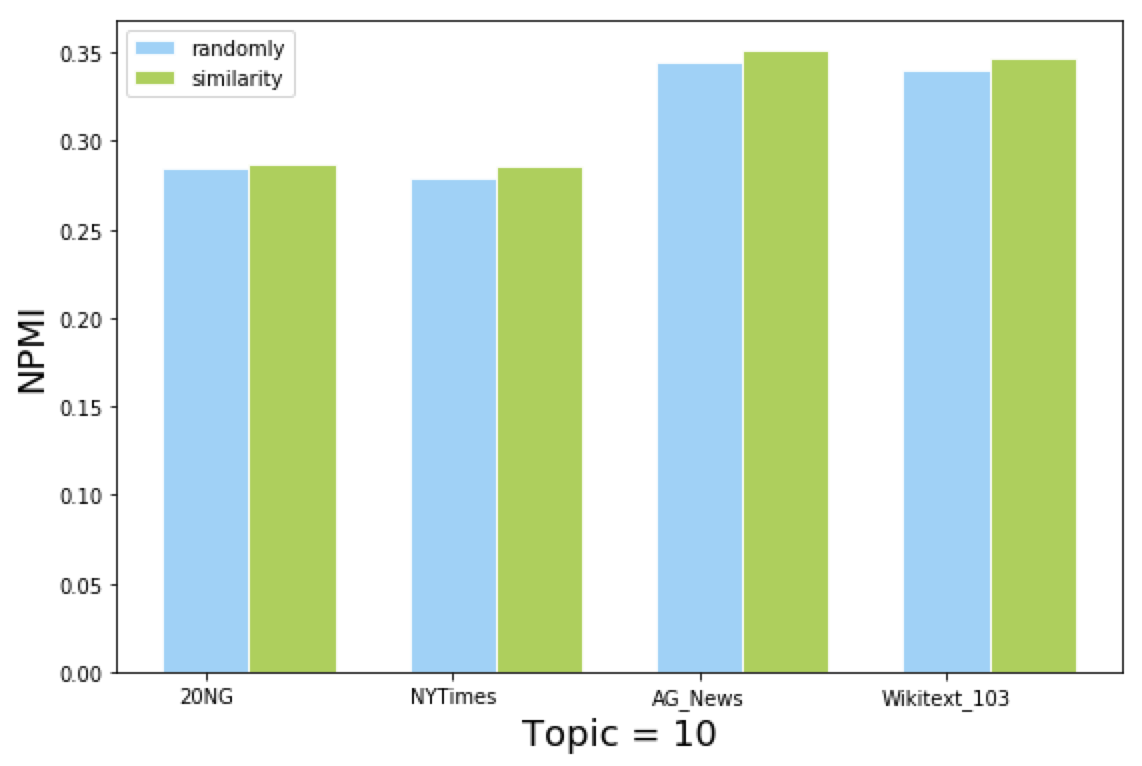}
\end{minipage}%
}%
\subfigure[Topic = 10\_TU.]{
\begin{minipage}[t]{0.45\linewidth}
\centering
\includegraphics[width=2in]{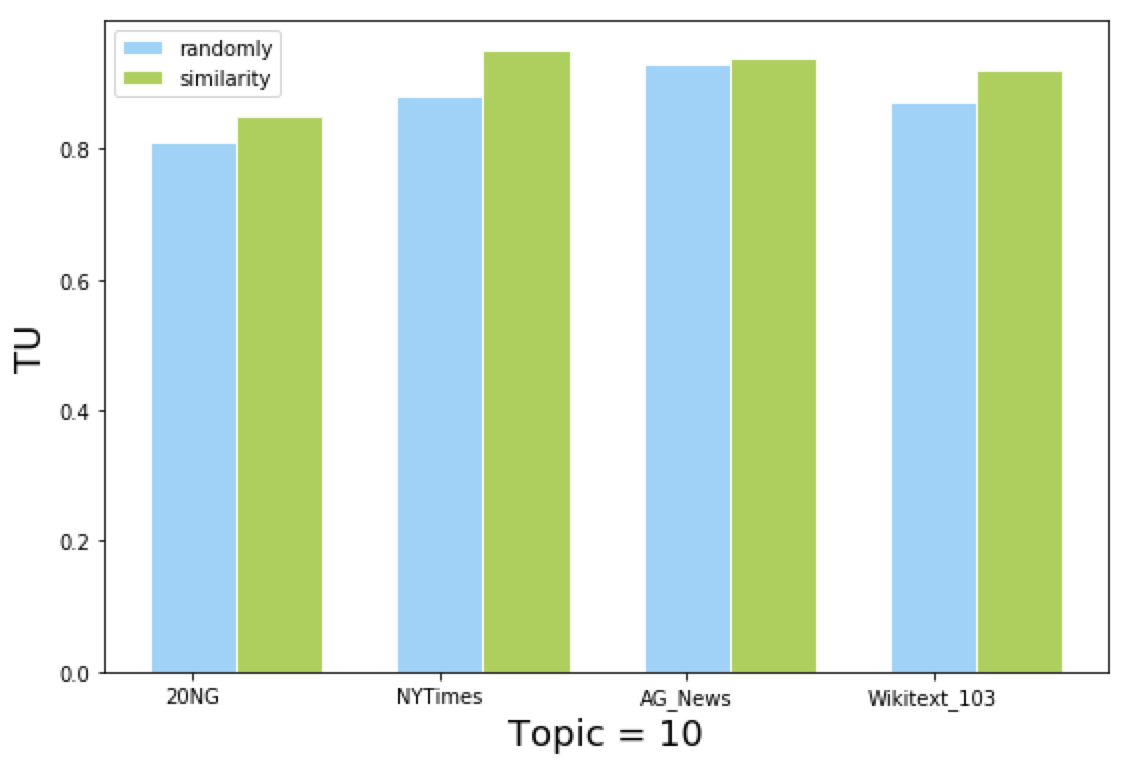}
\end{minipage}
}%
\caption{ NPMI and TU with different ways of negative sample choosing when Topic = 10.\label{fig:comp10} }
\end{figure}

\begin{figure}[H]
\centering
\subfigure[Topic = 20\_NPMI.]{
\begin{minipage}[t]{0.45\linewidth}
\centering
\includegraphics[width=2in]{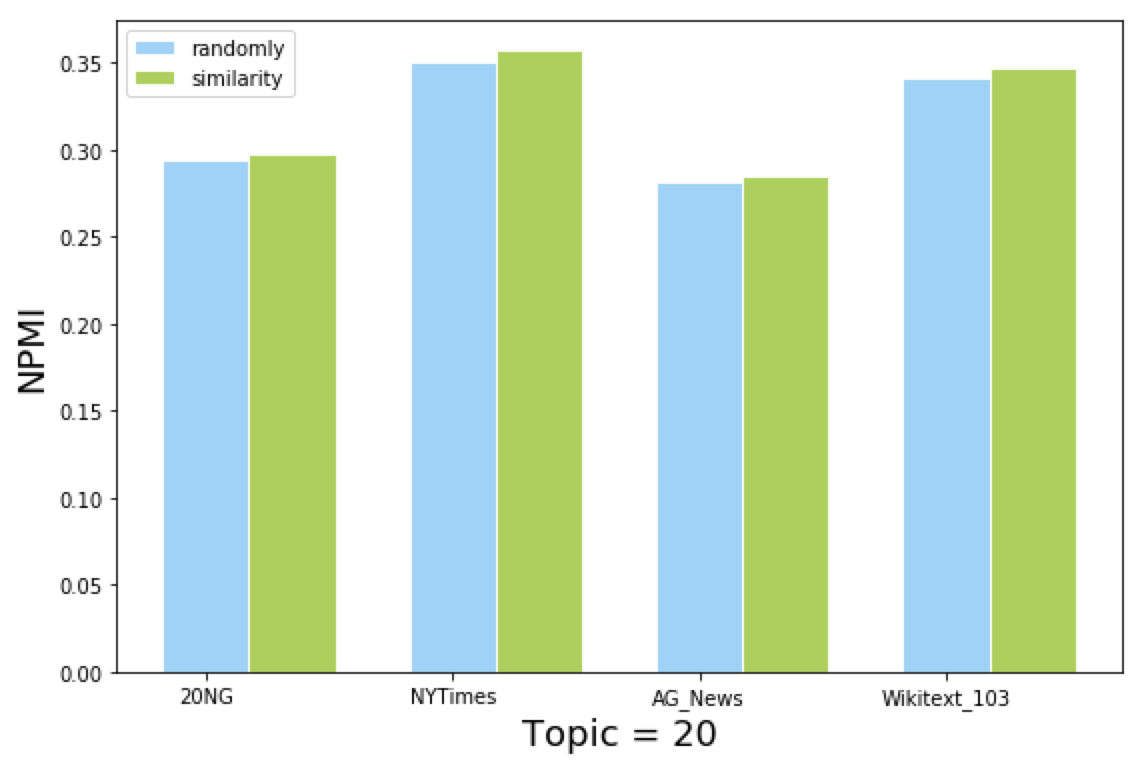}
\end{minipage}%
}%
\subfigure[Topic = 20\_TU.]{
\begin{minipage}[t]{0.45\linewidth}
\centering
\includegraphics[width=2in]{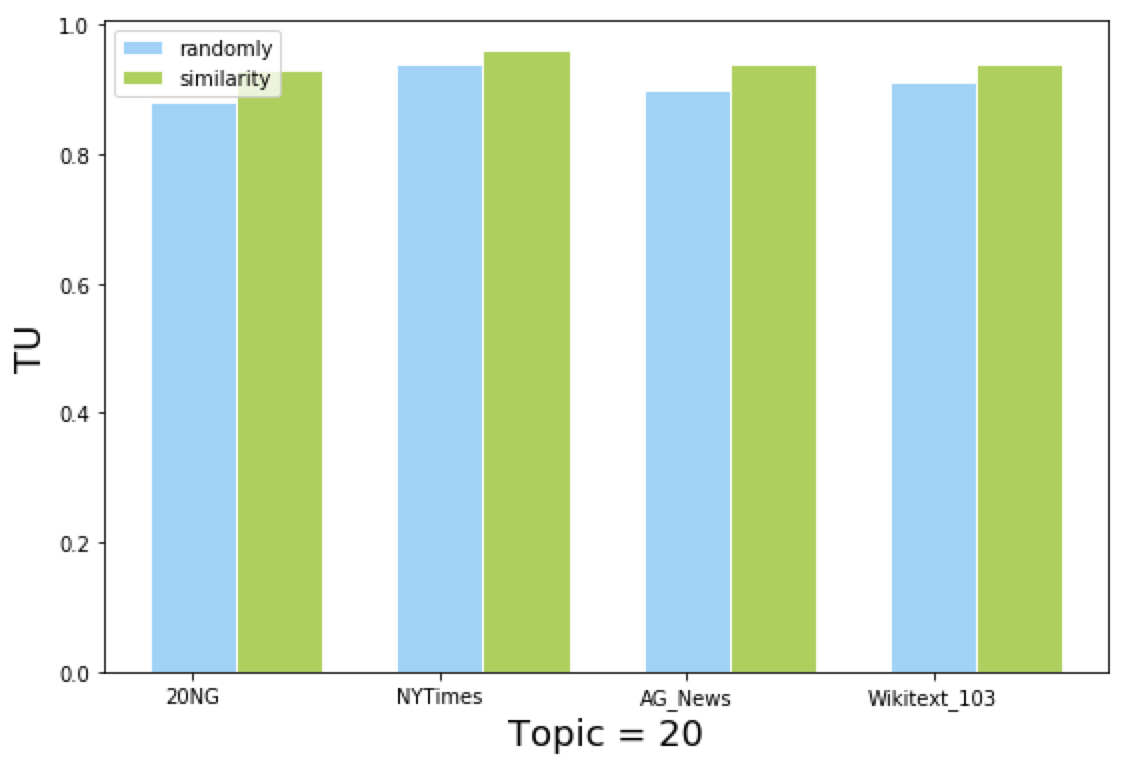}
\end{minipage}
}
\caption{ NPMI and TU with different ways of negative sample choosing when Topic = 20. \label{fig:comp20}}
\end{figure}

\begin{figure}[H]
\centering
\subfigure[Topic = 50\_NPMI.]{
\begin{minipage}[t]{0.45\linewidth}
\centering
\includegraphics[width=2in]{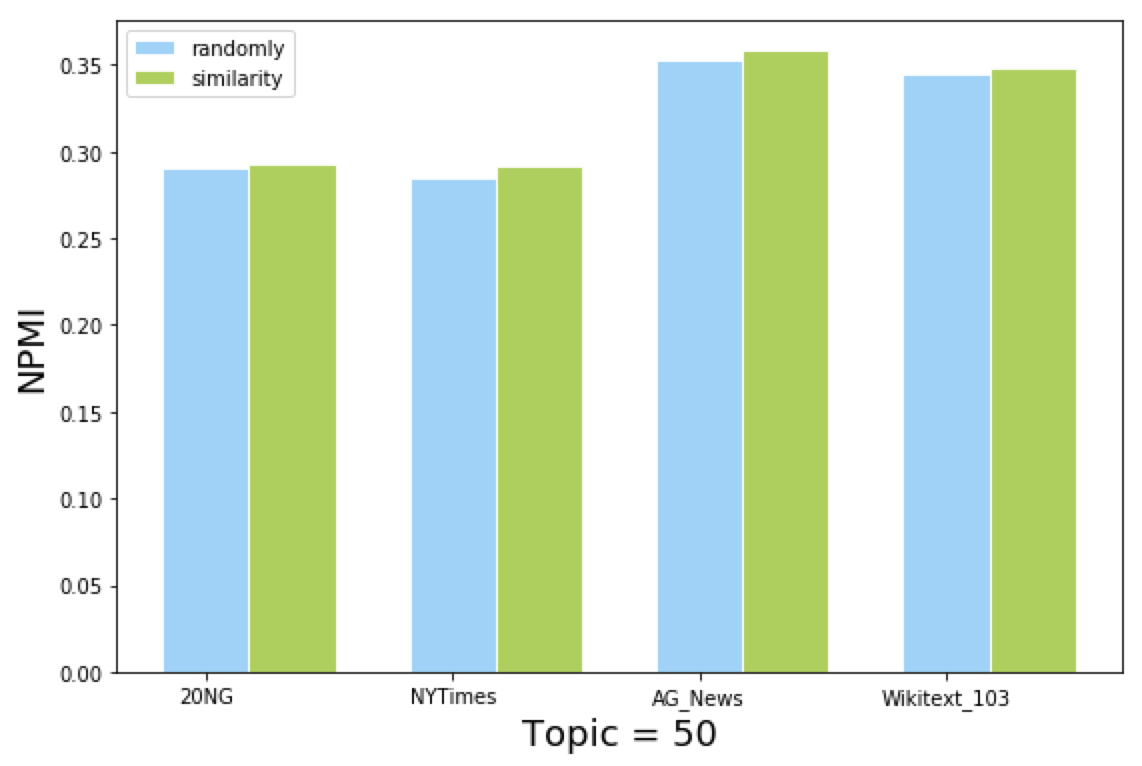}
\end{minipage}%
}
\subfigure[Topic = 50\_TU.]{
\begin{minipage}[t]{0.45\linewidth}
\centering
\includegraphics[width=2in]{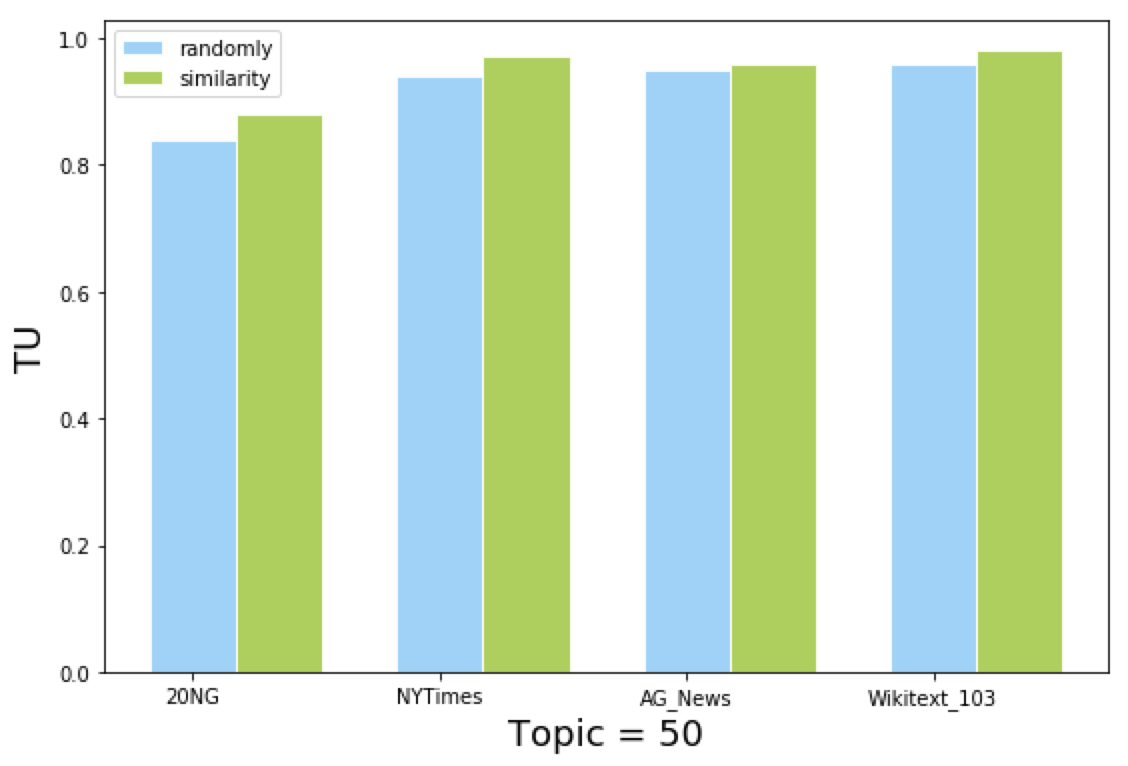}
\end{minipage}
}
\centering
\caption{ NPMI and TU with different ways of negative sample choosing when Topic = 50. \label{fig:comp50}}
\end{figure}

\begin{figure}[H]
\centering
\subfigure[Topic = 100\_NPMI.]{
\begin{minipage}[t]{0.45\linewidth}
\centering
\includegraphics[width=2in]{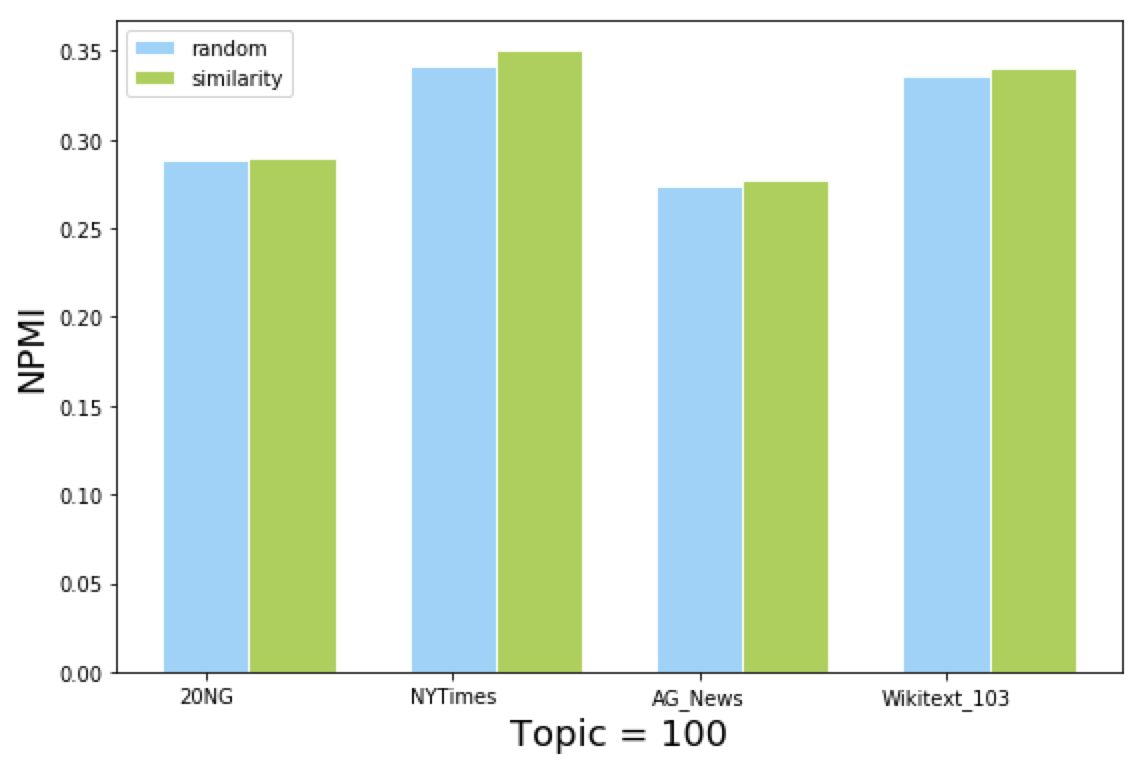}
\end{minipage}%
}
\subfigure[Topic = 100\_TU.]{
\begin{minipage}[t]{0.45\linewidth}
\centering
\includegraphics[width=2in]{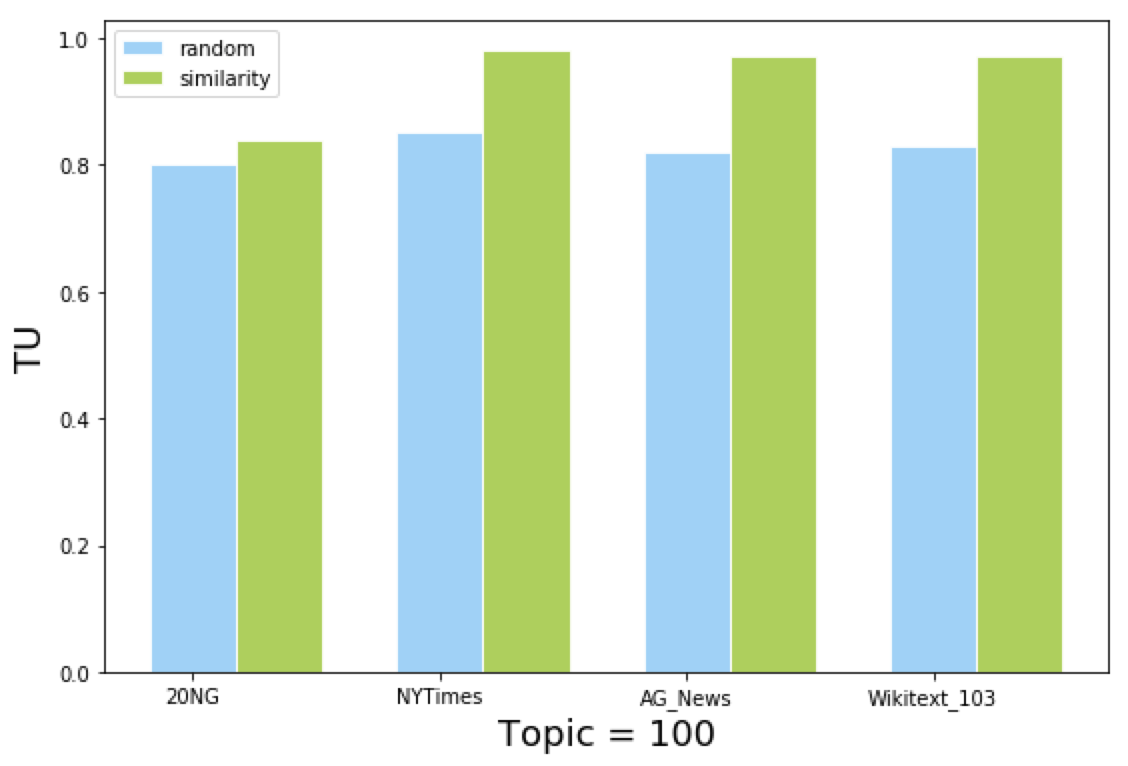}
\end{minipage}
}
\centering
\caption{ NPMI and TU with different ways of negative sample choosing when Topic = 100. \label{fig:comp100}}
\end{figure}

Table \ref{tab:npmitu} shows that NTM-DMIE performs better with the similarity-based selection strategy than with the random strategy. We can also see the trend in Figure \ref{fig:comp10}, Figure \ref{fig:comp20}, Figure \ref{fig:comp50} and Figure \ref{fig:comp100}. This makes intuitive sense as a randomly chosen ``negative'' example may not actually be sufficiently different from the given document, thus providing the discriminator with noisy training signals. 

\subsection{Qualitative Analysis}
\label{ssec:tv}
In order to more closely examine the accuracy of the topics and the corresponding keywords captured from each topic model, we compare NTM-DMIE with all the baseline models on the 20NG and NYTimes datasets, with the topic number set to 10. 
\begin{table}[H]
\centering
\caption{Topic-word example of Gaussain\_BAT and NTM-DMIE on 20Newsgroups dataset with topic = 10 }
\label{tab:20ngtopic}
\resizebox{\textwidth}{!}
{
\begin{tabular}{lcccccccccc}
\toprule
\multicolumn{2}{c}{Topic:Politics and Military}\\
\midrule
ProdLDA&gun people law problem government think kill death say drug\\
GSM&gun government tax pay law weapon rights people firearm president\\
NTM&people gun government law rights article drug death clinton kill\\
Scholar&gun article run people law government say death problem weapon\\
Gaussian\_BAT&gun problem people use government law article drug write kill\\
NTM-DMIE&gun government tax pay law weapon rights people firearm president\\
\bottomrule
\multicolumn{2}{c}{Topic:Transportation and daily life}\\
\midrule
ProdLDA&car bike ride get dod run write go put like\\
GSM&car bike problem use dod work ride driver set get\\
NTM&car bike problem get use ride dod driver work drive\\
Scholar&car bike drive dod ride get engine driver use buy\\
Gaussian\_BAT&car bike ball use dod work ride driver set get\\
NTM-DMIE&car bike drive ride engine dod buy sell like bmw\\
\bottomrule
\multicolumn{2}{c}{Topic:Security}\\
\midrule
ProdLDA&key chip use problem encryption bit system work clipper government\\
GSM&key chip encryption use run clipper system government problem bit\\
NTM&key chip encryption clipper get work government use system bit\\
Scholar&key use chip problem encryption work system set machine bit\\
Gaussian\_BAT&key chip use problem encryption bit system work clipper government\\
NTM-DMIE&key chip clipper encryption escrow nsa government system secure need\\
\bottomrule
\multicolumn{2}{c}{Topic:Communication}\\
\midrule
ProdLDA&drive card controller disk monitor thanks port pc driver system\\
GSM&card drive controller disk monitor use port controller driver window\\
NTM&drive card use problem disk work sale offer monitor machine\\
Scholar&drive card disk windows use run sale problem monitor pc\\
Gaussian\_BAT&drive card use disk monitor sale controller reply printer window\\
NTM-DMIE&card thanks please use window advance email port reply display\\
\bottomrule
\end{tabular}
}
\end{table}

\begin{table}[H]
\centering
\caption{Topic-word example of Gaussain\_BAT and NTM-DMIE on NYtimes dataset with topic = 10 }
\label{tab:nyttopic}
\resizebox{\textwidth}{!}{
\begin{tabular}{ll}
\toprule
\multicolumn{2}{c}{Topic:Business}\\
\midrule
ProdLDA&company money percent pay state cost bill industry buy chief\\
GSM&company executive business chief sell buy price pay share \\
NTM&company president executive business sell market sale chief share buy\\
Scholar&company percent market price business sell sale industry buy executive\\
Gaussian\_BAT&company computer system technology program number service price product information\\
NTM-DMIE&company percent market price business sell sale pay buy industry\\
\bottomrule
\multicolumn{2}{c}{Topic:Literature and Art}\\
\midrule
ProdLDA&music play art film book world write performance audience present\\
GSM&music art play book film write world television performance life\\
NTM&music program art book director write production company feature performance\\
Scholar&play music film performance movie audience young write book character\\
Gaussian\_BAT&play music write book life world art film character director\\
NTM-DMIE&music play art film movie book director write performance audience\\
\bottomrule
\multicolumn{2}{c}{Topic:Politics and Law}\\
\midrule
ProdLDA&case police charge official court law judge yesterday officer state\\
GSM&case police charge officer law man life feel official rule\\
NTM&law charge case court police judge recieve graduate official rule\\
Scholar&state law issue official vote public case court member judge\\
Gaussian\_BAT&case police charge official court law worker judge public officer\\
NTM-DMIE&case law official court state charge issue judge rule police \\
\bottomrule
\multicolumn{2}{c}{Topic:People and Life}\\
\midrule
ProdLDA& man life feel thing tell ask woman friend son student\\
GSM&life man woman young thing write son love tell friend\\
NTM&man police woman life death son daughter family child kill\\
Scholar&man life father mother family student child graduate director school\\
Gaussian\_BAT&man life woman father mother child young family feel friend\\
NTM-DMIE&shcool child father family mrs son mother student graduate daughter\\
\bottomrule
\end{tabular}
}
\end{table}

Table \ref{tab:20ngtopic} shows that nearly all the models can extract the four topics, namely Politics and Military; Transportation and daily life; Security; and Communication. However, NTM-DMIE can mine more keywords that are more closely related to the specific topics than the other models, while other baselines may mine some words that do not belong to the corresponding topic. For example, in topic Politics and Military, Gaussain\_BAT captures \textbf{"article"} and \textbf{"write"}, which are not so closely related to politics or military while NTM-DMIE can capture words like \textbf{"firearm"} and \textbf{"president"} which are closely related to the topic. Table \ref{tab:nyttopic} also shows that nearly all the models can extract the four topics, namely Business; Literature and Art; Politics and Law; and People and Life. However, similarly, NTM-DMIE performs better than the other five baselines.

\section{Conclusion}
\label{ssec:conc}
In this paper, we have proposed a framework to incorporate deep mutual information into neural topic modeling. Our framework maximizes the mutual information between the input documents and their latent topic representations. We capture mutual information on the global and local levels to preserve the rich information of the documents and words into their topic representations. 
A discriminator is also employed to discriminate a document from its negative examples for learning robust topic representations. Experiments on four public datasets show that our model outperforms state-of-the-art neural topic models on the metrics of topic coherence and topic uniqueness. A further experiment on text clustering demonstrates the quality of the learned topics in downstream tasks. In future work, we will investigate self-supervised learning approaches to extend our model. 

\section{Acknowledgements}
\label{ssec:ack}
We would like to thank the reviewers for their comments, which helped improve this paper considerably. This work was supported in part by the Research Foundation for Advanced Talents of Nanjing University of Posts and Telecommunications under Grants NY218118 and NY219104, in part by an open project of the State Key Laboratory of Smart Grid Protection and Control, Nari Group Corporation, under Grant SGNR0000KJJS2007626, and in part by the Jiangsu Project of Social Development under Grant BE2020713.

\bibliography{mybibfile}

\end{document}